\documentclass{article}
\usepackage[margin=0.75in]{geometry}
\usepackage{libertine}
\usepackage[style=english]{csquotes}
\usepackage[style=ieee, url=false, doi=false]{biblatex}

\usepackage{hyperref}
\hypersetup{
	colorlinks = true,
	citecolor = {black},
	linkcolor = black
}

\usepackage{graphics} 
\usepackage{epsfig} 
\usepackage{amsmath} 
\usepackage{amssymb}  
\usepackage{bm}
\usepackage{subfig}
\usepackage{array}
\usepackage{algorithm, algorithmic}

\usepackage{graphicx}
\usepackage{float} 

\title{Towards Learning Controllable Representations \\ of Physical Systems}
\author{Kevin Haninger, Raul Vicente Garcia, J\"org Kr\"uger \footnote{Corresponding email: \texttt{kevin.haninger@ipk.fraunhofer.de}. Authors affiliated with the Fraunhofer Institute f{\"u}r Produktionsanlagen und Konstruktionstechnik (IPK), Department of Automation.}%
}

\begin{document}
\maketitle
\begin{abstract}
	Learned representations of dynamical systems reduce dimensionality, potentially supporting downstream reinforcement learning (RL). However, no established methods predict a representation's suitability for control and evaluation is largely done via downstream RL performance, slowing representation design. Towards a principled evaluation of representations for control, we consider the relationship between the true state and the corresponding representations, proposing that ideally each representation corresponds to a unique true state. This motivates two metrics: temporal smoothness and high mutual information between true state/representation. These metrics are related to established representation objectives, and studied on Lagrangian systems where true state, information requirements, and statistical properties of the state can be formalized for a broad class of systems. These metrics are shown to predict reinforcement learning performance in a simulated peg-in-hole task when comparing variants of autoencoder-based representations.
\end{abstract}

\section{Introduction}
Representation learning aims extract `useful' information for downstream learning tasks \cite{bengio2013}, typically by an unsupervised training objective. For  example, the variational autoencoder (VAE) \cite{kingma2013} learns an encoder and decoder which approximately reconstruct the original data. The VAE and many other representation learning techniques were developed for large corpora of text, image, or audio data; three distinctions can be raised in image-based control in robotics:  (i) robotics typically has a single downstream task -- control, (ii) data are not i.i.d. but correlated according to the robot/environment dynamics, (iii) there are proprioceptive sensors on the robot, e.g. position and force, which can provide additional information. 

A wide range of representation learning methods have been proposed. VAEs have been applied to pixel control in robotics to compress images independently (i.e. with no dynamics) \cite{finn2015, vecerik2018, yarats2020}. A dynamic model can also be simultaneously learned for a time series of images, with either linear \cite{watter2015, banijamali2018} or nonlinear dynamics \cite{hafner2019, lee2020a}. For these methods, the training loss remains pixel reconstruction, where the dynamics provides additional terms to the ELBO training loss. Contrastive learning - temporal \cite{sermanet2018} or via data augmentation \cite{srinivas2020} - has also been applied to robotics problems. A representation loss can be purely unsupervised \cite{lee2020a}, or include a reinforcement learning (RL) loss \cite{srinivas2020}. This breadth of representation learning methods, and the reliance on empirical comparisons of RL performance, which introduces further noise to the comparison, makes design and evaluation of the representation loss and architecture difficult.
\begin{figure}[t]
	\centering
	\includegraphics[width=.25\columnwidth]{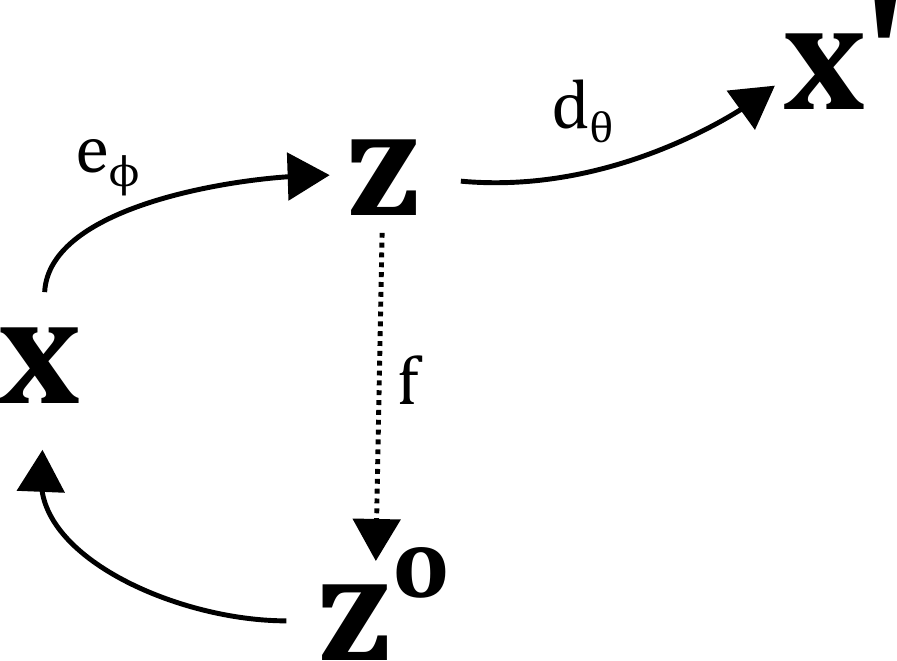} 	\protect\caption{An autoencoder applied to a dynamic system, where true state trajectory $z^o$ produces observations $x$, which are compressed by an encoder $e$ to a latent representation $z$. Here, it is proposed that at each time step, $z_n$ should correspond to a unique $z^o_n$, i.e. $\exists f,\,\, z^o_n=f(z_n)$. \label{ground_truth}}
\end{figure}

Evaluation metrics for general representation learning focus on disentanglement \cite{locatello2018, chen2018b}, whereby meaningful `factors of variation' should appear independently in the representation. For most dynamic systems, there are well-motivated factors of variation, here referred to as true state and denoted $z^o$. However, disentanglement in robotics merits reconsideration - for example, a representation which independently presents joint positions is likely less useful than one where end-effector coordinates are independent.

The objective of uniqueness is here proposed; that each sample of the representation $z_n$ correspond to a unique true state $z^o_n$. If a representation corresponds to multiple true states, a control policy over the representation cannot articulate any differences in action which may be required. While perfect information is likely not necessary for any given control task, this objective may help decouple representation and reinforcement learning as the representation objective is not task-dependent.

A general statement of the true state can be made by Lagrangian mechanics for rigid-body and quasi-rigid systems: the generalized position and velocity of the robot and its environment. This true state is controllable - it provides sufficient information for control \cite[\S~6.5]{bloch2003}. Lagrangian mechanics also provides a coordinate-free description of dynamics and informs how additional sensors (force, robot position) relate to the true state, which will be here used to derive statistical properties of the true state and motivate the encoder architecture.

This paper operationalizes the uniqueness principle in two ways: mutual information between true state and representation, and smoothness. Mutual information $MI(z^o; z)$ slightly differs from the classical objective of $MI(x;z)$ \cite{phuong2018}, and because $z^o$ is low-dimensional the former can be estimated with techniques such as MINE \cite{belghazi2018}. Smoothness with respect to $z^o$ can be estimated based on statistical properties of the true state, allowing formal statements of when temporal smoothness should be expected on Lagrangian systems. These metrics are shown to predict downstream reinforcement learning performance on a simulated peg in hole task. 

\subsection{Related Work}
VAEs are not the only method for representation learning in robotics, but they are popular. In many cases, a standard VAE is trained on images without any temporal information \cite{lange2012, finn2015, yarats2020}, even if a dynamic model is later fit over the latent $z$ \cite{ha2018}.  Other approaches simultaneously learn the representation and dynamics over the representation. Restricting these dynamics to be locally linear \cite{watter2015, banijamali2018, zhang2018} allows the use of closed-form control policies. More general nonlinear dynamics can be considered \cite{lee2020a, hafner2019}. These approaches are variational, where the dynamics adds additional terms to the ELBO loss. This improves the RL performance \cite{lee2020a} at the cost of complexity. Representations can also be learned by time-contrastive learning \cite{dwibedi2018, sermanet2018} or contrastive learning via data augmentation \cite{srinivas2020}. As autoencoders with carefully chosen training data, losses, and regularization achieve good results \cite{yarats2020} and remain common in robotic representation learning, this paper focuses on VAEs. 

The aforementioned techniques focus on the compression and learning from pixels alone. In robotics applications, the integration of existing sensors (force, position) improve performance in representation learning for dexterous, contact-rich  manipulation \cite{lee2020c, triyonoputro2019, luo2019a}. Auxiliary losses such as success classification \cite{vecerik2018} have also been employed in contact tasks, and shown to improve the ensuing performance of an RL agent. 

The temporal smoothness of a representation is a classical objective in representation learning \cite{wiskott2002}, and has been ad-hoc proposed as a loss in robotic representation learning \cite{finn2015, jonschkowski2017}. This is also related to (single view) time-contrastive learning \cite{sermanet2018}, and has been proposed as a representation evaluation metric \cite{raffin2018}. Here, it is rigorously motivated, and conditions under which it would be expected on Lagrangian systems found. 

Evaluation methods for general representations are also proposed, many being posed from `disentanglement' \cite{bengio2013}, a recent review can be found in \cite{locatello2018}. Methods for evaluating representations for continuous control have been proposed \cite{raffin2018, antonova2020}, with metrics of coherence (the spread between adjacent representations when projected to true state), correlation between true state and representation, and reconstruction error when regressing from latent to true state. Here, these metrics are augmented with mutual information and empirically connected with RL task performance. 

\subsubsection*{Notation}
Unless otherwise noted, $z=[z_0, \dots, z_N]$, a trajectory of random continuous variables where $z_n$ is defined over sample space $Z$, similarly $x$ and $z^o$. Standard notation for expectation $\mathbb{E}$ and normal distribution $\mathcal{N}(\mu,\Sigma)$, with mean $\mu$ and covariance $\Sigma$ are used. 

\section{Controllability of physical systems}
Traditionally, controllability is a system property - whether there exists an input trajectory $\tau=[\tau_1,\dots,\tau_N]$ to reach an arbitrary state for some finite $N$ \cite{aastrom2010}. We refer to a representation $z$ as controllable if there exists a control policy $\tau_n=\pi(z_n)$ which achieves similar objectives.

Lagrangian systems, which describe continuous-time dynamics of rigid and quasi-rigid physical systems, have a true state of $z^{o}(t)=[q(t),\dot{q}(t)]$, with configuration $q(t)\in Q$ (generalized position) and its time derivative $\dot{q}(t)$. The configuration $q(t)$ determines stored energy at time $t$ and any holonomic constraints $h\left(q(t)\right)=0$, which can describe contact conditions between the robot and its environment. The configuration can also describe aspects of the environment which vary between task iterations - e.g. the relative position of a door hinge. Robot position is typically measured (a subset of $Q$) and any force measurements, either joint torque or 6-DOF force/torque, are functions of $z^o$. 

A general constrained Lagrangian system can be written as
\begin{equation}
\label{equations_of_motion}
\begin{aligned}
M\left(q\right)\ddot{q}+C\left(q,\dot{q}\right) & =\frac{\partial h\left(q\right)}{\partial q}^{T}\lambda+B\left(q\right)\tau\\
h\left(q\right) & =0 
\end{aligned}
\end{equation}
with holonomic constraints $h:Q\rightarrow\mathbb{R}^{m}$, inertia matrix $M$, Coriolis and gravitational terms $C$, input joint torque $\tau$ modulated by an input matrix $B$, Lagrangian multipliers $\lambda\in \mathbb{R}^m$ which enforce the constraint, and where the time argument $(t)$ has been suppressed. The initial condition is distributed as $q(t=0)\sim p(q_0)$.

For any a smooth manifold $N=\{q,\dot{q}|\dot{q}=0,\,n(q)=0\}$, there exists a smooth control policy $\tau(t)=\pi^o(q,\dot{q})$ such that $N$ is locally asymptotically stable --- points near the manifold converge to $N$ over time and do not leave (see, e.g. \cite[Theorem~3]{bloch1992}, or  \cite[\S 6.5]{bloch2003}). We assume sufficient fast sample rate $T_s$ such that similar performance can be achieved by the discrete time policy $\tau_n=\pi^o(z^o_n)$, where $z^o_n=[q(nT_s), \dot{q}(nT_s)]$. As the objective in most robotic tasks is a relative positioning of the robot and environmental bodies with a final relative velocity of $0$, we consider $z^o$ as a controllable representation. 

Furthermore, if there exists a smooth surjective function $f: Z\rightarrow Z^o$ where $z_n=f(z^o_n)$, then $z$ is also controllable; the true state control policy $\pi^o$ can be directly employed as $\tau_n=\pi^o(f(z_n))$. The existence of $f$ implies that each $z_n$ corresponds to a unique true state $z^o_n$, thus motivating the uniqueness principle.

\section{Operationalizing uniqueness}

This section introduces two ways to operationalize uniqueness for representation learning; mutual information and smoothness.

\subsection{Mutual information}

Consider the maximization of mutual information between $z_n$ and $z^{o}_n$ 
\[
\max_{\phi}\sum_{n=0}^NI_{\phi}\left(z_n;z^{o}_n\right).
\]
The $n$ subscript on $z_n$, $z^o_n$, $x_n$ will be suppressed for the remainder of subsection III.A. Mutual information $I\left(z;z^{o}\right)=\iint p\left(z,z^{o}\right)\log\frac{p\left(z,z^{o}\right)}{p\left(z\right)p\left(z^{o}\right)}dzdz^o$ is a symmetric non-negative measure of dependence between the two random variables, which is zero at independence and $\phi$ are parameters of the encoder. The mutual information can be interpreted as the reduction in uncertainty of $z^{o}$ when $z$ is known, i.e.
\[
I_{\phi}\left(z;z^{o}\right)=H\left(z^{o}\right)-H_{\phi}\left(z^o|z\right)
\]
where $H\left(z^{o}\right)=\int p\left(z^{o}\right)\ln p\left(z^{o}\right)dz^o$ is the differential entropy of $z^o$. $H\left(z^{o}\right)$ is not a function of the representation, thus maximizing $I_{\phi}\left(z,z^{o}\right)$ over $\phi$ corresponds to minimizing conditional entropy $H_{\phi}\left(z^o|z\right)$; i.e. given $z$, remaining uncertainty in $z^o$ is low. 

A function $\hat{f}$ can also be directly fit to minimize $\mathbb{E}\Vert z-\hat{f}(z^o)\Vert$, showing the (approximate) existence of a function from representation to true state. This has been used to validate a representation \cite{hafner2019, yarats2020}, and is compared here in the experimental validation in Section \ref{experiments}. Mutual information has the advantage of connecting with broader goals in representation learning.

\subsubsection*{Bounds on mutual information}
A bound on $I\left(z;z^{o}\right)$ can be found, rewriting $I\left(z;x,z^{o}\right)$ with the chain rule $I(z;x,z^{o})=I(z;z^o)+I(z;x|z^o)=I(z;x)+I(z;z^o|x)$ \cite{cover1999} and noting the Markov chain $z^{o}\rightarrow x\rightarrow z$ implies $I\left(z;z^{o}|x\right)=0$, giving:
\begin{eqnarray}
I_\phi\left(z;z^{o}\right) & = &I_\phi\left(z;x\right)-I_\phi\left(z,x|z^{o}\right) \label{mi_decomp}\\
 & \leq &I_\phi\left(z;x\right) \label{mi_decomp_ineq}
\end{eqnarray}
where \eqref{mi_decomp_ineq} is by nonnegativity of mutual information. $I_\phi\left(z;x\right)$, the mutual information between encoder input and latent, is an upper bound on $I(z;z^o)$ and is common representation learning \cite{barber2003, poole2019}. However, as $z^o$ is typically lower dimension than $x$, estimating $I(z^o;z)$ is easier with established techniques. 

The objective of maximizing \eqref{mi_decomp_ineq} can be connected to the rate-distortion viewpoint of the $\beta$-VAE \cite{higgins2017} developed in \cite{alemi2017}, where it is shown that $I_\phi(z;x)\leq R$, where $R$ is the encoding rate of the encoder. The training loss of the $\beta$-VAE can be rewritten as
\begin{eqnarray*}
\mathcal{L}(\phi,\theta) &=& \mathbb{E}_{z\sim e_\phi\left(z|x\right)}\left[d_\theta\left(x|z\right)\right]+\beta KL\left(e_\phi\left(z|x\right)||p\left(z\right)\right) \\
&=& D+\beta R \nonumber
\end{eqnarray*}
where $D=\mathbb{E}_{z\sim e_\phi\left(z|x\right)}\left[d_\theta\left(x|z\right)\right]$ is the distortion, the reconstruction error. A smaller $\beta$ reduces the penalty on encoder rate, raising an upper bound on $I(x;z)$. A smaller $\beta$ improves resulting control performance in the experimental results here (Section \ref{experiments}) and elsewhere \cite{yarats2020}. Other adaptations to the VAE have been proposed to increase $I\left(x;z\right)$ \cite{phuong2018a,zhao2018b}.


\subsection{Smoothness}
This section formalizes an argument that under certain conditions, temporal smoothness in representation $z$ is desired, e.g. that $\mathbb{E}\Vert z_{n+1}-z_n \Vert$ should be small. This objective is often proposed ad-hoc \cite{bengio2013}, appearing in `slow features' \cite{wiskott2002}, time contrastive learning \cite{sermanet2018}, and has also been used as an auxiliary loss in robotic representation learning \cite{finn2015, jonschkowski2017}. By making a formal argument here, the conditions under which temporal smoothness implies representation uniqueness are made explicit.

Returning to the objective of a representation where $\exists f$ such that $z^o_n=f(z_n)$: if $z_n, z^{o}_n\in \mathbb{R}^k$, denote a deterministic observation and encoding pipeline as $z_n=g\left(z^{o}_n\right)$.  This assumes a memoryless observation and encoder process, and a deterministic encoder. Note that while a stochastic encoder with noise in $z$ is useful for regularizing a representation during training \cite{higgins2017}, during RL the maximum likelihood $z$ is typically passed to the controller \cite{yarats2020}.  Denoting the Jacobian $J_g = \frac{\partial g}{\partial z^{o}_n}$, recall the differential entropy is related as
\begin{eqnarray}
H\left(z_n\right)-\mathbb{E}\ln\left|J_g\right|\leq H\left(z^{o}_n\right)\label{entropy_trans_var}
\end{eqnarray}
where $|\cdot|$ is the absolute value of the determinant and equality obtains if and only if $g$ is invertible (i.e. $\exists f=g^{-1}$ such that $z^o_n=f(z_n)$) \cite[\S 14]{papoulis2002}.
\begin{figure}
	\centering
	\includegraphics[width=.5\columnwidth]{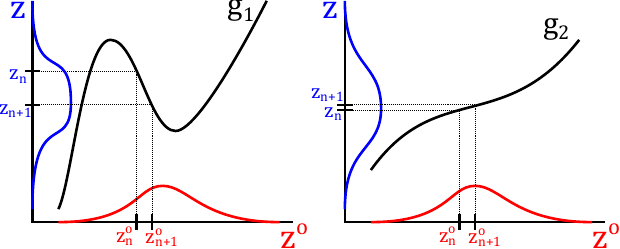} 
	\protect\caption{Two transformations from true state $z^o_n$ to latent $z_n$, constructed such that $H(g_1(z^o_n))=H(g_2(z^o_n))$ but $E\ln |\frac{dg_1}{dz^o_n}|\neq \mathbb{E}\ln |\frac{dg_2}{dz^o_n}|$. Function $g$ is invertible when $E \ln |\frac{dg}{dz^o_n}| $ is reduced, which can be estimated from successive time steps $n$ and $n+1$. \label{smoothness}}
\end{figure}

As $H\left(z^{o}_n\right)$ is not changed by the representation, equality in \eqref{entropy_trans_var} is approached by maximizing $H(z_n)-\mathbb{E}\ln|J_g|$. That is, a good representation has high entropy (a wide distribution), but varies slowly with respect to changes in true state. If $H(z_n)$ is constrained by the training loss (e.g. the KL divergence between $p(z|x)$ and a unit normal distribution in VAEs \cite{kingma2013}), minimizing $ \mathbb{E}\ln |J_g| $ approaches equality in \eqref{entropy_trans_var}. 
 
$|J_g|$ can be approximated from the time-series properties of $z^{o}$. Consider an initial $z_n$ and $z^o_n$ where $z_n=g\left(z^{o}_n\right)$, and define perturbations $\delta_{z}$, $\delta_{z^{o}}$. A first-order Taylor approximation gives $\delta_{z}\approx J_{g}\delta_{z^{o}}$. If $\delta_{z^{o}}\sim\mathcal{N}\left(0,\alpha I\right)$, $\alpha>0$, then $\delta_{z}\sim\mathcal{N}\left(0,\alpha J_{g}J_{g}^{T}\right)$. Denote eigenvalues of $J_g$ of $\lambda_i$, then $\ln | J_g | = \sum_{i=1}^k \ln \lambda_i$, $\ln \mathbb{E}\Vert\delta_z\Vert_2^2 = \ln \sum_{i=1}^k \lambda_i^2+\ln \alpha$, and 
\begin{align}
\frac{k}{2}\left(\ln \mathbb{E}\left\Vert \delta_{z}\right\Vert _{2}^{2}-\ln\alpha k\right) & \geq \ln\left|J_{g}\right| \label{pert_jac_ls}
\end{align}
by the log sum inequality, where equality is attained if and only if $\lambda_i=\lambda_j$, $\forall i,j$. Thus, when $H(z)$ is constrained and $\delta_{z^o}$ is normal with zero mean with unit covariance, minimizing $\mathbb{E}\left\Vert \delta_{z}\right\Vert$ brings \eqref{entropy_trans_var} closer to equality. Minimizing $\mathbb{E}\ln |J_g|$ alone can be achieved by simply scaling the coordinates of $z$ -- smoothness requires additional constraints on the representation to be useful.

\subsubsection*{Temporal Smoothness in Lagrangian systems}
Denote temporal first difference $\delta_{z^{o}}^1=z_{n+1}^{o}-z_{n}^{o}$, and second difference $\delta_{z^o}^2 = z_{n+1}^{o}-2z_{n}^{o}+z_{n-1}^o$, which have both been used to measure smoothness \cite{jonschkowski2017, finn2015}. Taking dynamics of \eqref{equations_of_motion} where holonomic constraint $h(q)=0$ has been eliminated, the explicit Euler approximation is

\begin{eqnarray}
\delta_{z^o}^1 =\left[\begin{array}{c}
q_{n+1}-q_{n}\\
\dot{q}_{n+1}-\dot{q}_{n}
\end{array}\right] \approx T_{s}\left[\begin{array}{c}
\dot{q}_{n}\\
M^{-1}_n\left(C_n+B_n\tau_n\right)
\end{array}\right] \label{ltv_dyn}
\end{eqnarray}
where $T_s$ is the sample time, $M_n = M(q_n)$, similarly $B_n, C_n$. Considering \eqref{ltv_dyn} as a linear time-varying system (i.e. ignoring the nonlinearity from system matrices), if $C_n$ is compensated, initial velocity $\dot{q}_0=0$, and $\tau_{n}\sim\mathcal{N}\left(\pi_n,\Sigma_n\right)$, 
\begin{equation}
\delta_{z^o}^1\sim \mathcal{N}\left(\left[\begin{array}{c} \sum_{i=0}^n \tilde{B}_i\pi_i \\ \tilde{B}_n\pi_n \end{array}\right],\left[\begin{array}{cc}
\sum_{i=0}^{n}\tilde{B}_{i}\Sigma_i\tilde{B}_i^T\\
 & \tilde{B}_{n}\Sigma_n\tilde{B}_n^T
\end{array}\right]\right)
\label{first_dif}
\end{equation}
where $\tilde{B}_n = T_sM^{-1}(q_n)B(q_n)$. When $\pi_n=0$ (i.e. random exploration), $\delta_{z^o}^1$ is a zero mean Gaussian. Furthermore, if eigenvalues $\lambda_i(\tilde{B}_n\Sigma_n)=1$ (i.e. higher magnitude exploration noise in `slower' degrees of freedom), position change $z^o_{n+1}-z^o_n$ has unit covariance. 

If $\tilde{B}_n\approx \tilde{B}_{n-1}$, and $\pi_{n+1}-\pi_n = \delta_\pi$
\begin{equation}
\delta_{z^o}^2\sim\mathcal{N}\left(\left[\begin{array}{c} T_s \tilde{B}_n\pi_{n-1} \\ \tilde{B}_n\mathbb{E}[\delta_\pi] \end{array}\right],\left[\begin{array}{cc}
T_s \tilde{B}_n\Sigma_n\tilde{B}_n^T\\
& \tilde{B}_n(\Sigma_n \mathtt{+} \Sigma_{n\mathtt{-}1}+\mathbb{E}[\delta_\pi\delta_\pi^T])\tilde{B}_n^T
\end{array}\right]\right).
\label{second_difs}
\end{equation}
Taking a first order approximation to the policy, $\delta_\pi \approx \frac{d\pi}{dz^o}\delta_{z^{o}}^1$. When $\pi$ is smooth; these terms will be small. In comparison with \eqref{first_dif}, $\delta_{z^o}^2$ reduces the bias in positions by a factor of at least $T_s$. Additionally; the covariance of $\delta_{z^o}^2$ does not grow with $n$, providing a more stationary distribution.

\section{Validation \label{experiments}}
\begin{figure}
	\centering
	\subfloat[\label{architecture}]{\includegraphics[width=.45\columnwidth]{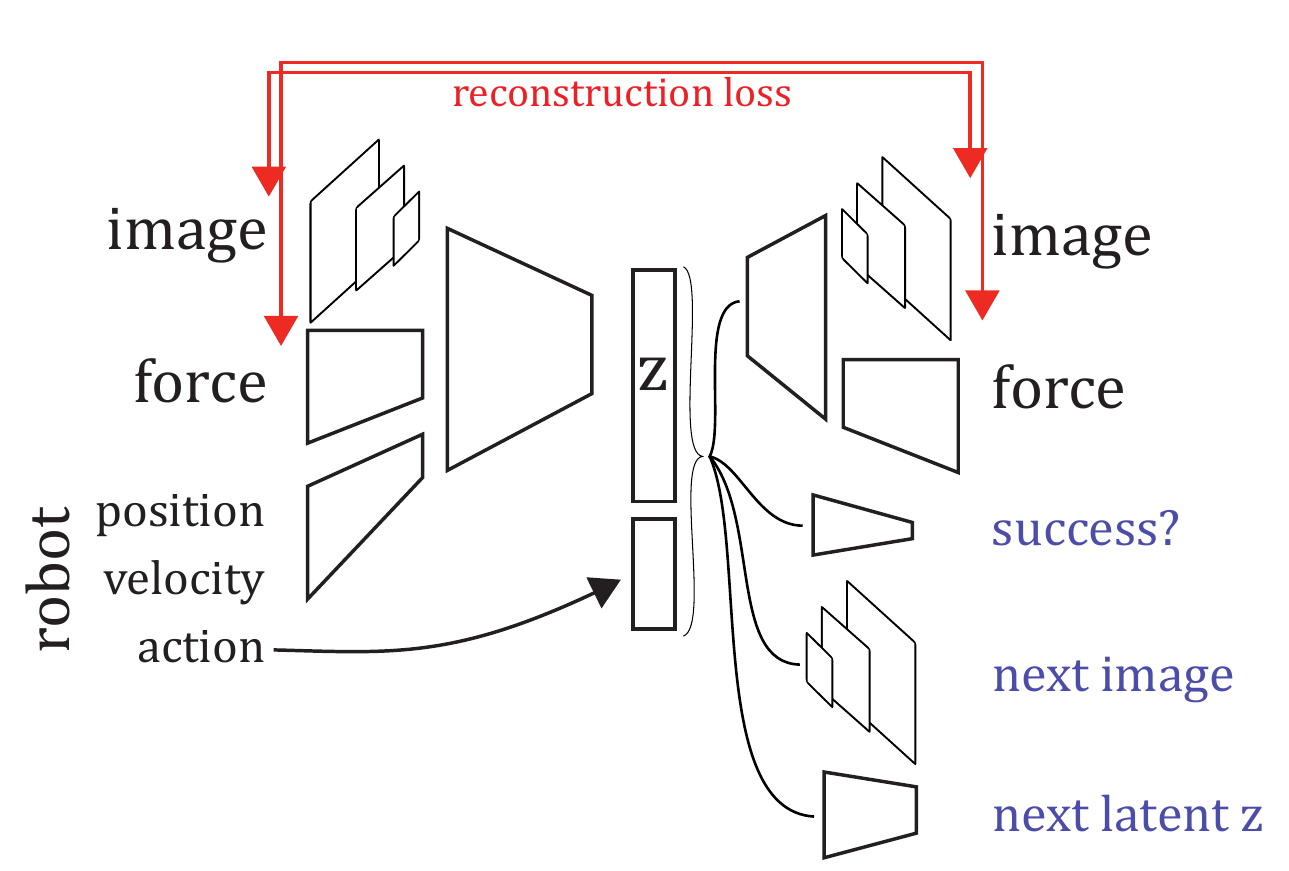}} 
	\hfill
	\subfloat[ \label{data_coll}] {\includegraphics[width=.27\columnwidth]{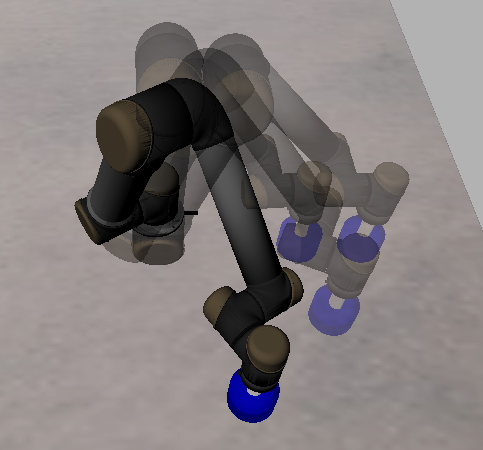}} 
	\hfill
	\subfloat[]{\begin{tabular}[b]{c}
		\includegraphics[width=.125\columnwidth]{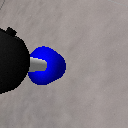} \\  \includegraphics[width=.125	\columnwidth]{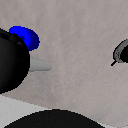}
		\end{tabular}}
	\hfill
	\hfill
	\caption{Autoencoder architecture (a), with auxiliary losses in blue, and overview of the peg-in-hole simulation (b). Hand camera images (c) are collected from random exploration from the goal, over 25 different environment configurations, and used to train the encoders. \label{peg_in_hole}}
\end{figure}
These experiments are conducted in a simulated peg-in-hole task where the hole position varies between iterations, a wrist camera provides 64x64 RGB images, and a 6-DOF force/torque sensor is mounted at the wrist. A dataset is collected with backwards exploration from the goal position as seen in Figure \ref{data_coll}, for 25 different hole positions (total 7500 samples). Several autoencoders are trained on this dataset, all of which are based on the classical VAE:
\begin{itemize}
	\item {\tt VAE}, reconstruct image and force measurements \cite{kingma2013}
	\item {\tt success}, add a success classifier \cite{vecerik2018}
	\item {\tt pred$_Z$}, add a loss for predicting the next latent state, based on current latent state and action
	\item {\tt pred$_{IM}$}, penalizes reconstruction of the true next image \cite{lee2020a, vanhoof2016, antonova2020}
	\item {\tt smooth}, which adds a smoothing loss $\Vert z_{t+1}-z_t\Vert $ \cite{jonschkowski2017}. 
\end{itemize}
The baseline autoencoder is implemented as seen in Figure \ref{architecture}, where direct measurements from the robot of position, velocity, low-pass filtered force measurements, and torque input are processed by a modality-specific MLP, then fused with fully-connected layers to produce the distribution of $z$.  The latent $z$, concatenated with torque input, is used to decode the original image and force.  The convolutional stack for the images increases from 32 to 128 filters, filter size 4, stride 2/1, with ReLU activations.  The force and proprioception encoders have one hidden layer of size 16, with a fusion encoder has a hidden layer of 64 units, fully-connected to mean and log standard deviation for $z$.  The decoder inverts this stack.

The success classifier contains 2 FC layers followed by softmax (success labels are identified by hand). The next image predictor has an additional FC layer before the deconvolution stack. The next latent predictor uses a hidden layer of size 64, ReLU activations for mean and log standard deviation of $z_{n+1}$. The loss weights used for training are reconstruction $1e-4$, KL weight is $10$ normally, and $0.1$ with low KL encoders. The success weight is $0.1$, smoothing weight $500$, prediction of next $z$ weight $1.0$ and the prediction of next image weight $1e-4$. On the low KL smoothing, the smoothing weight was reduced to $0.1$.

After encoder training, two model-free RL agents are trained: SAC \cite{haarnoja2018b} and DDPG \cite{lillicrap2015a}, based on the code associated with \cite{yarats2020}. The state given to the agent is the latent state (maximum likelihood, i.e. $\mu_z$ not a sample),  augmented with the robot state. The agents are trained over  10 seeds each, with actor and critic architecture of two hidden layers of size 96 or 128, trained with Adam at learning rates $0.003$ or $0.001$, replay buffer of 3e6, batch size of 256, and soft-update $\tau$ of $0.005$. Noise tuning was used on SAC, and DDPG had constant exploration noise of magnitude $0.25$ the maximum input. RL agents are  trained for 150k environment steps and evaluated for 20k steps.  A video of the resulting performance is available at: \url{https://youtu.be/adHXi58u9qw} . A second dataset collects representation trajectories produced  by a trained full-state SAC agent, to allow comparison between the representations on-policy in some of the following analysis.

\subsection{Mutual information and performance}
\begin{figure}
	\centering
	\subfloat[Mutual information vs evaluation performance \label{evaluation_perf}]{\includegraphics[width=.55\columnwidth]{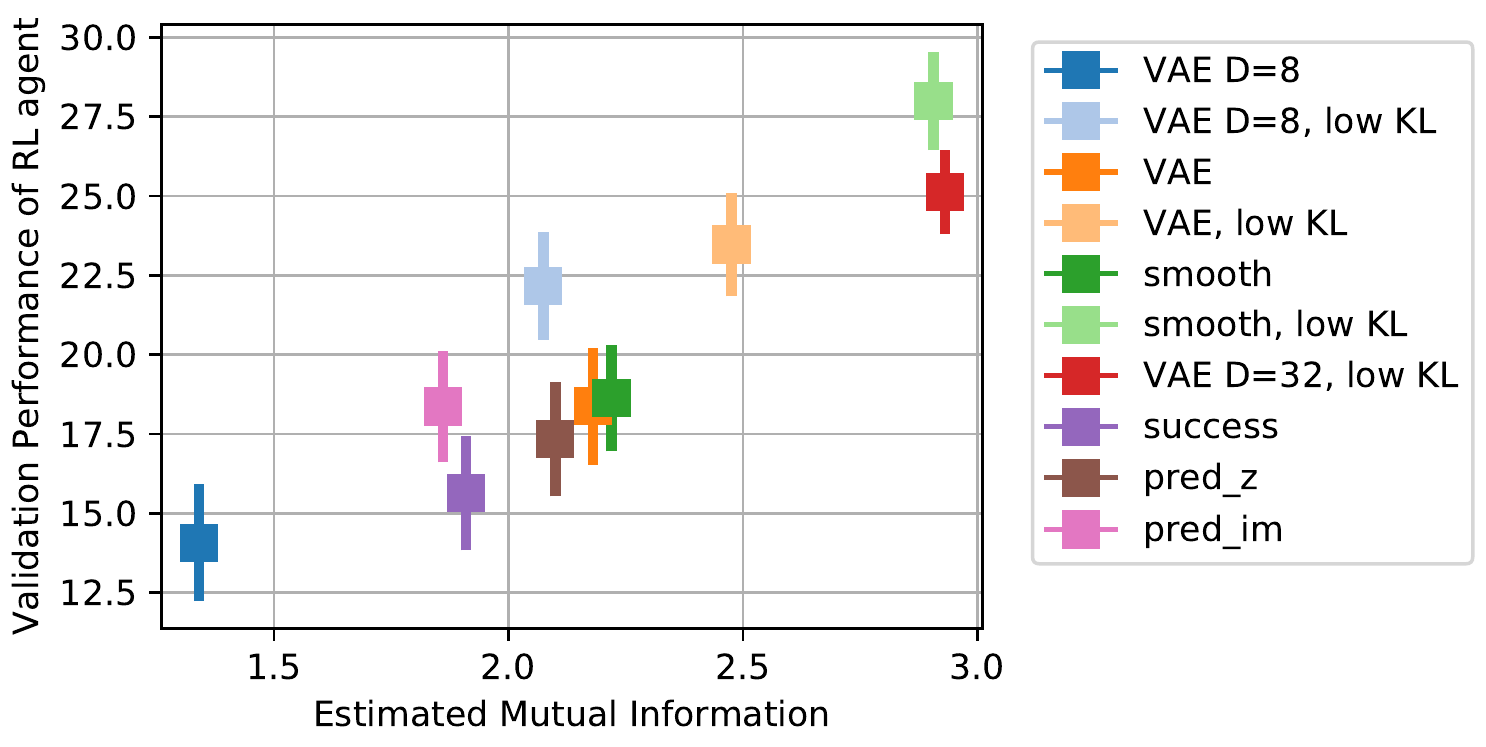}}\hspace{.01\columnwidth} 
	\hfill
	\subfloat[Training performance of encoders trained with standard and low KL weights \label{training_perf}] {\includegraphics[width=.35\columnwidth]{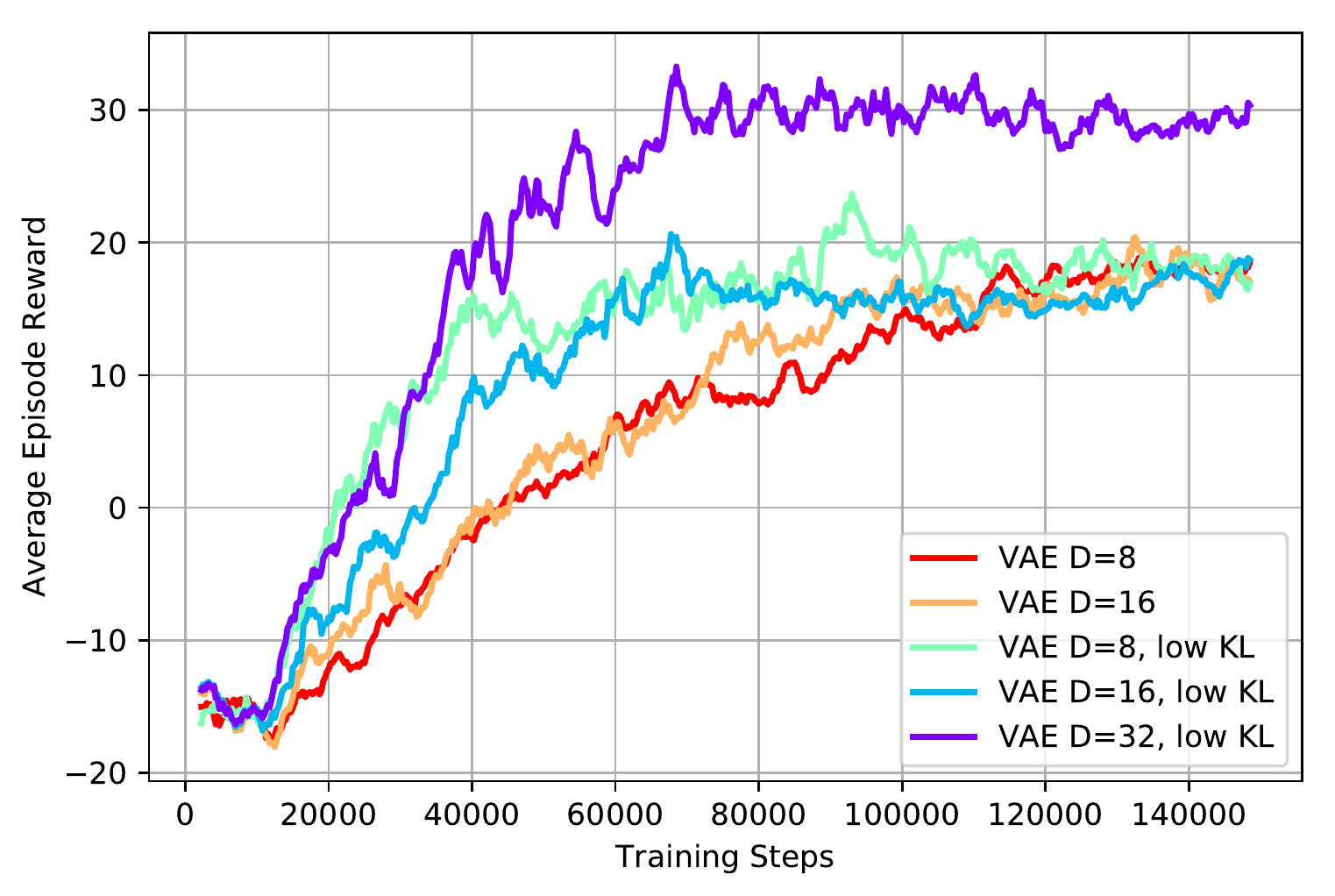}}  	
	\caption{Training and evaluation performance of the encoders, latent dimension $D=16$ unless otherwise noted. Higher mutual information between latent and true state correlates with RL training speed and end performance.}
\end{figure}
The mutual information between $z_n$ and $z^o_n$ is estimated with MINE \cite{belghazi2018}, where the true state is the position/velocity of the robot, and the planar position of the hole. MINE is implemented with linear/ReLU layers with 64 hidden units, optimized with Adam at a learning rate of $5e-5$, batch size of 128. In Figure \ref{evaluation_perf}, a strong correlation between estimated mutual information and RL performance can be seen over all tested encoders.  The performance of the auxiliary training losses (\texttt{pred$_Z$, pred$_{IM}$}) is largely similar; there is not significant variation in performance or mutual information, while \texttt{success} slightly reduces both performance and mutual information in this example. The smoothness penalty increases the mutual information and performance in low KL case, but has more limited impact in the standard case.

Reducing the KL weight in the training loss significantly improves RL performance, as noted in existing work \cite{yarats2020}, and mutual information. This is also indicated by the $\beta$-VAE analysis in \cite{alemi2017}, which suggests decreasing the KL weight increases the mutual information $I(x;z)$ of the encoder. The tuning of the KL weight and the encoder architecture is substantially more important for performance in this task. As seen in Figure \ref{training_perf}, the encoders with a lower KL weight also train faster than their counterparts, while the dimension of the latent space does uniformly impact training speed. 

\subsection{Regression to True State}
\begin{figure}[h]
	\centering
	\subfloat[L2 Regression error vs RL performance  \label{reg_vs_perf}]{\includegraphics[width=.55\columnwidth]{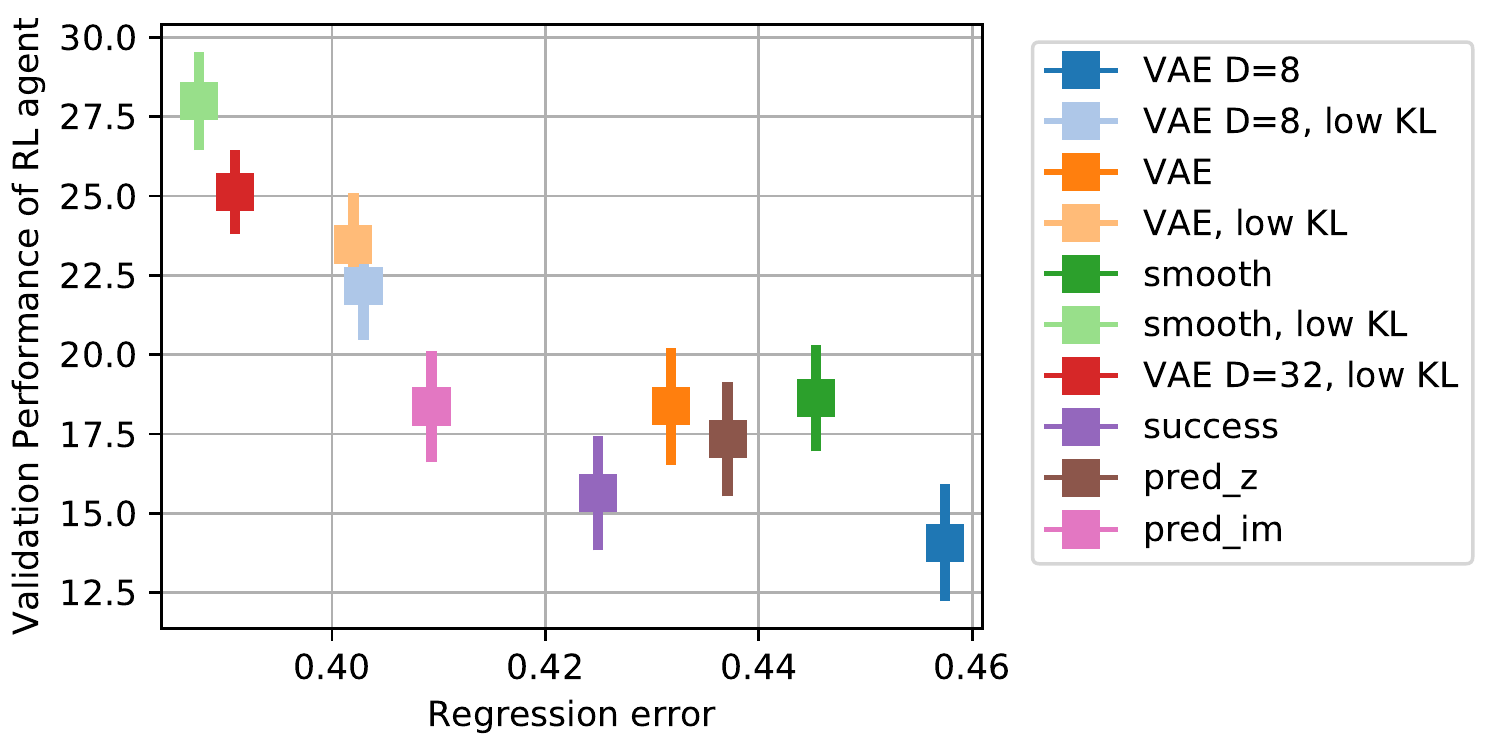}}  
	\hfill
	\subfloat[L2 Regression error vs mutual info \label{reg_vs_mi}] {\includegraphics[width=.35\columnwidth]{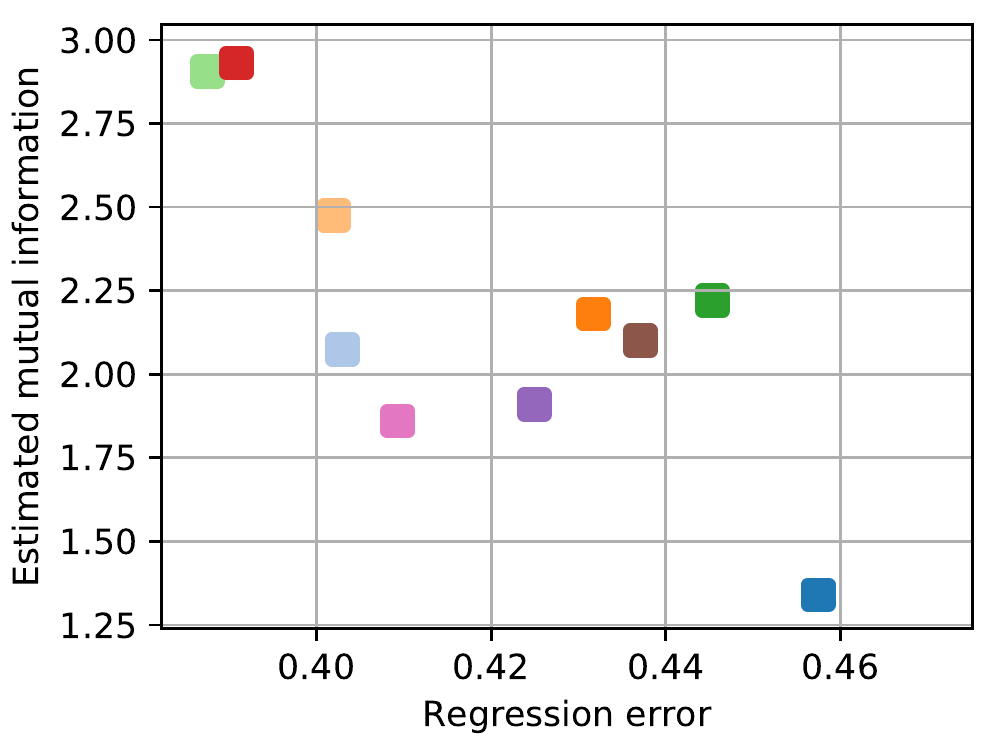}}  	 		
	\caption{Regression error in fitting a function from latent to true state, $z^o = \hat{f}(z)$. Lower regression error correlates with higher RL task performance and higher mutual information, validating regression as a validation tool for a representation.}
\end{figure}

Regression from latent to the true state has also been employed as a validation in representation learning \cite{hafner2019, yarats2020}, and proposed as a metric for representation quality \cite{raffin2018, antonova2020}. Here, the residual regression error is compared with RL performance and estimated mutual information in Figure \ref{reg_vs_perf} and \ref{reg_vs_mi}, respectively. The regression results use FC layers with ReLU activations, two hidden layers of 64 units, trained to minimize the L2 error on the combined random and on policy datasets, with 30\% of the data held out for the validation, validation score reported in the figure.

The L2 regression error also has a strong correlation with RL performance, suggesting that it also serves well as a metric for representation quality. Accordingly, a good correlation can be seen between the regression error and mutual information. Further experiments in other environments are required to make any general conclusions regarding the relative merits of regression error and mutual information as metrics.

\subsection{Encoder Smoothness}
\begin{figure}[h]
	\centering
	\subfloat[Average latent distance at various time offsets. High KL top, low KL bottom. \label{latent_distances}]{\includegraphics[width=.45\columnwidth]{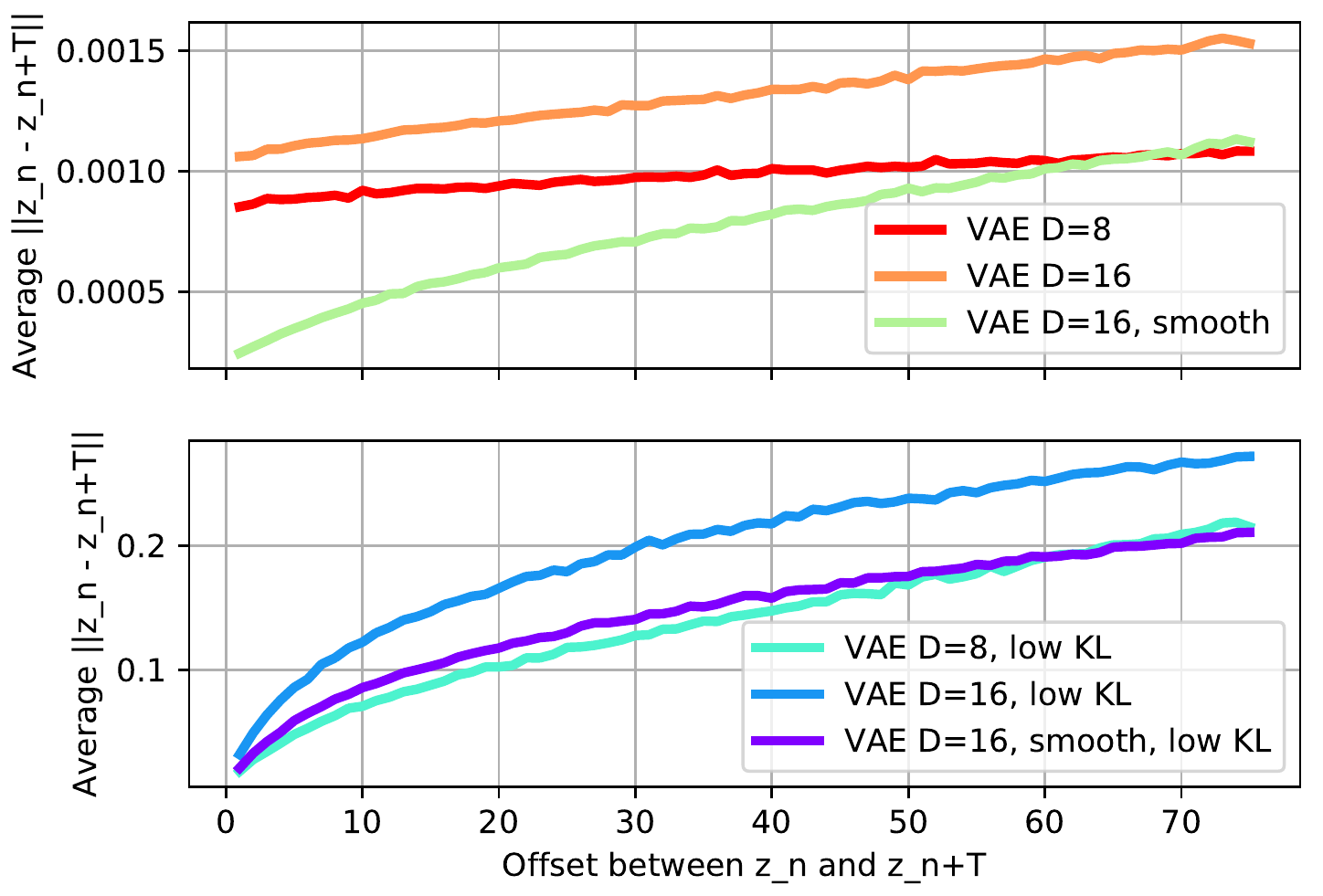}} 
	\hfill 
	\subfloat[Histogram of encoder smoothness $\frac{\Vert z_n-z_{n-1}\Vert}{\Vert z^o_n-z^o_{n-1}\Vert}$\label{encoder_smoothness}] {\includegraphics[width=.45\columnwidth]{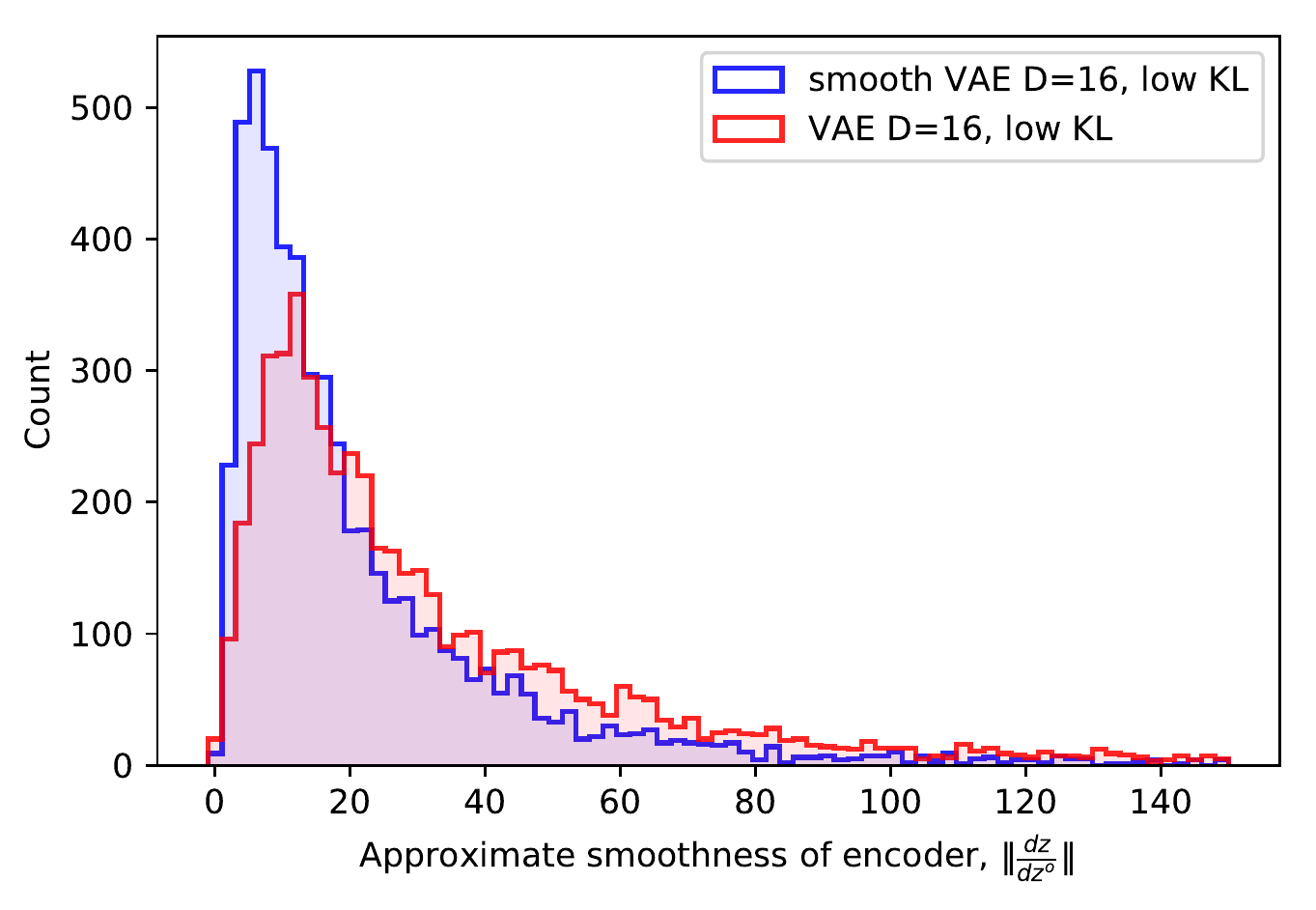}}  	 		
	\caption{Temporal and smoothness properties of selected representations. All representations are closer in latent space when temporally close, but the change is more pronounced on representations which have higher performance. In (b), the smoothing loss does result in a smaller average change in latent for a given change in true state.}
\end{figure}

The assumptions and efficacy of the smoothing loss will be examined on the same dataset.  First, the average distance between latent state at various time offset $T$, $\mathbb{E}\Vert z_n - z_{n+T}\Vert$, is shown in Figure \ref{latent_distances}. Latent states which are temporally close are spatially closer for all encoders, but to varying degrees.  The lower KL losses have latent representations which `grow apart' more, and also had better RL performance. The VAE produces time-contrastive representations, even absent temporal information. 

The smoothness of the encoding pipeline $z=g(z^o)$ is then measured as $\mathbb{E}\Vert z_n-z_{n+1}\Vert / \Vert z^o_n-z^o_{n+1} \Vert$, as shown in Figure \ref{encoder_smoothness} for the on-policy data (i.e. all representation trajectories are from the same true-state trajectory). The smoothed encoder has a lower average slope, changing less for small changes in true state. 
\begin{figure}[h]
	\centering 
	\subfloat[Distribution of first and second difference of true state $z^o$ under random agent and on policy, where bias predicted in (7) and (8) can be seen. \label{true_state_dist}]{\includegraphics[width=.45\columnwidth]{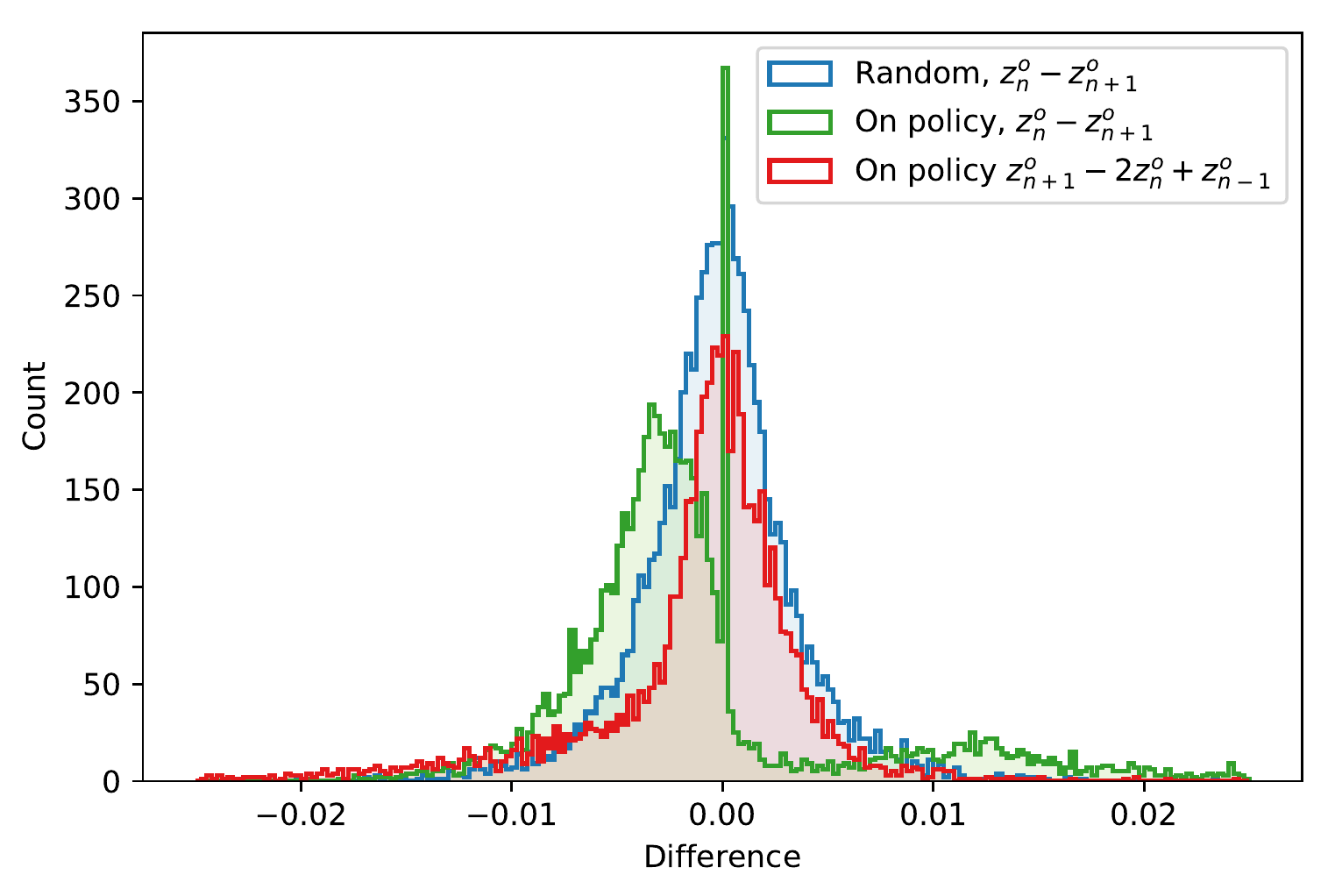}}  
	\hfill 
	\subfloat[Representations which are high entropy and smooth, i.e. large $H(z)-E \ln | J_g |$, have higher RL task performance, as predicted by the smoothness objective in \eqref{pert_jac_ls}. \label{smoothness_vs_perf}]{\includegraphics[width=.45\columnwidth]{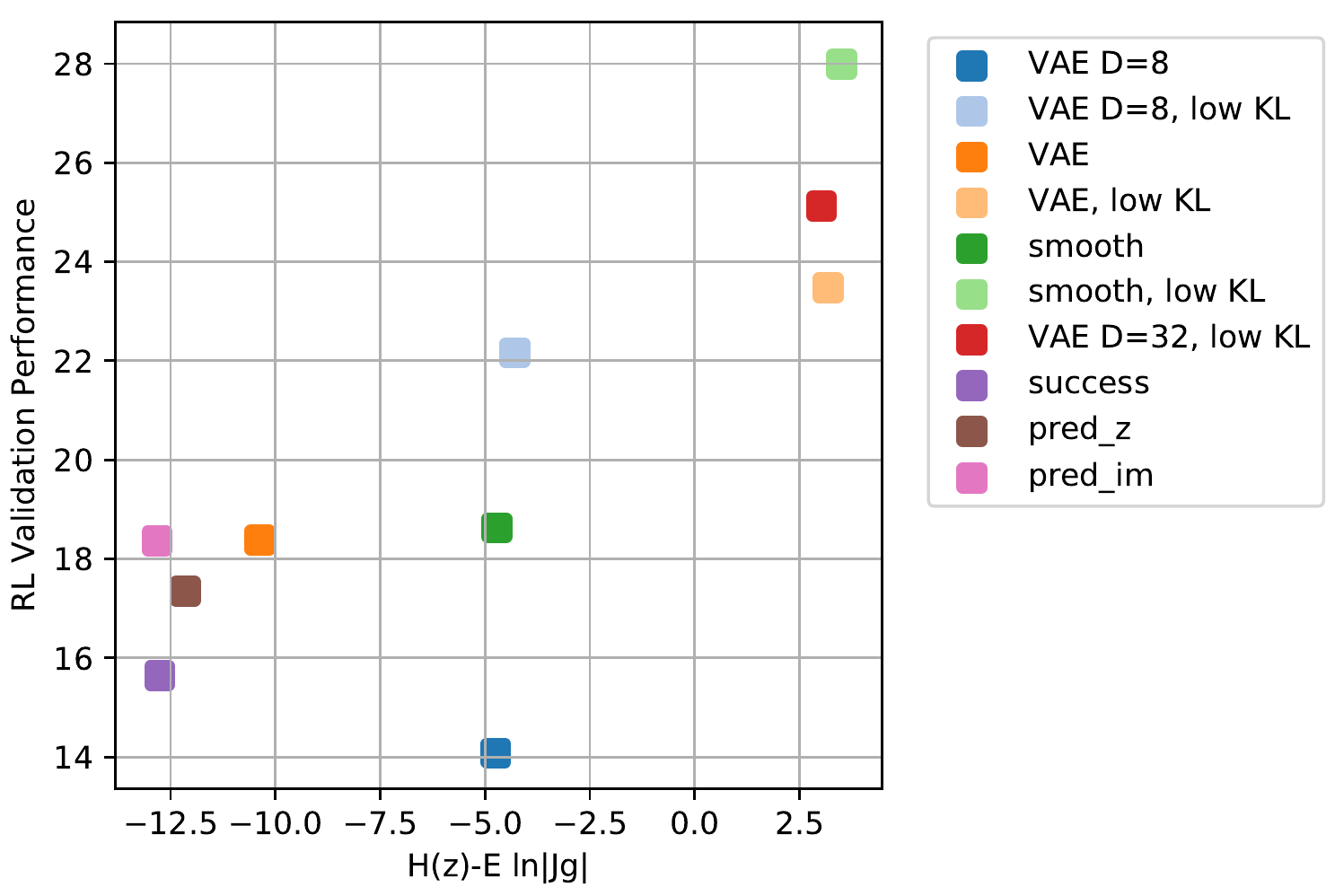}}	
	\caption{True state distribution and smoothness vs. RL performance} 		
\end{figure}

In \eqref{pert_jac_ls}, it was assumed that the first difference $z^o_{n}-z^o_{n+1}$ is normally distributed. The distribution of an element and the sum of the true state can be seen in Figure \ref{true_state_dist}, in both random and on-policy data. The first difference on the random agent (blue) is approximately normal, but the first difference on policy (green) is bimodal - the trained agent is moving towards the objective. However, the second difference on policy (red) removes this bias and is again approximately normal. 

The objective of approaching equality in \eqref{entropy_trans_var} can be investigated, estimating $\mathbb{E} \ln | J_g |$ with the upper bound from \eqref{pert_jac_ls} of $ \frac{n}{2}\left(\ln \mathbb{E} \Vert z_n-z_{n-1} \Vert^2_2 - \ln k\alpha\right)$, where $\alpha=0.12$ is used and $k$ is the dimension of the representation.  The entropy is estimated with a kNN estimator \cite{kraskov2004}, allowing comparison between $H(z)-\mathbb{E}\ln|J_g|$ and performance, shown in Figure \ref{smoothness_vs_perf}. Representations which are high entropy and smooth (small $|J_g|$) approach equality in $H(z)-\mathbb{E}\ln |J_g| \leq H(z_o)$, where equality implies there is a function $f$ such that $z_o=f(z)$. The relationship seen in Figure \ref{smoothness_vs_perf} is not as strong as mutual information or regression, but suggest the objective in \eqref{pert_jac_ls} is sensible. 

\section{Conclusion}
Towards a connection between properties of a representation and its suitability for control, this paper proposed the objective of a unique representation; where each representation $z$ corresponds to a unique true state $z^o$. This goal motivates two metrics: mutual information between $z, z^o$ and the temporal smoothness of $z$. The proposed metrics show good predictive power for model-free reinforcement learning performance. While the true state is typically unknown, the first or second difference of the true state are normal, and zero-mean under certain conditions. 

Further experimentation is needed to validate the initial results here on a wider range of robotics tasks, and on a wider range of representations (e.g. contrastive learning). However, the initial results can inform the tuning of hyperparameters and architecture of representation learning, which can greatly assist in hardware experiments with relatively expensive data collection. 


\printbibliography

\appendix
\section{Appendix}
\subsection{Estimating Mutual Information, Smoothness, Regression}
MINE \cite{belghazi2018} is used for the mutual information estimates reported in the text, based on implementation \cite{kim}.  These were compared with k-nearest neighbor approaches \cite{kraskov2004}, where many of the same trends emerge but the KNN methods struggled with low-entropy random variables. MINE provided high-variance estimates, ameliorated with large batch sizes, a lower learning rate, and a small moving average constant (0.001) for the unbiased estimator.  The network consist of linear/ReLU layers with 64 hidden units, Adam is used for optimization, with a learning rate of $5e-5$, batch size of 128. 

The regression results use FC layers with ReLU activations, two hidden layers of 64 units, trained to minimize the L2 error on the combined random and on policy datasets, with 30\% of the data held out for the validation, validation score reported in the figure.

The smoothness is estimated as in \eqref{pert_jac_ls} with $n$ as the latent state dimension, and $\alpha=0.12$, estimated from histograms of $z^o_n - z^o_{n+1}$. Entropy $H(z)$ is estimated with the implementation \cite{versteeg} of the kNN estimator \cite{kraskov2004}, with $k=5$. 

\subsection{Encoder}
\begin{figure}
	\centering
	\subfloat[Samples at $n$] {\includegraphics[width=.45\columnwidth]{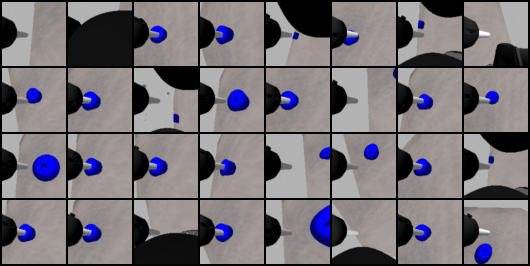}} \hspace{.01\columnwidth}
	\subfloat[Reconstructed samples at $n$]{\includegraphics[width=.45\columnwidth]{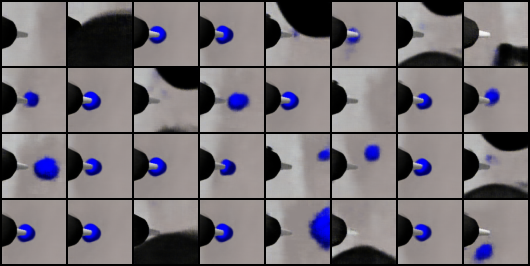}} \\ 
	\subfloat[Successor samples at $n+1$] {\includegraphics[width=.45\columnwidth]{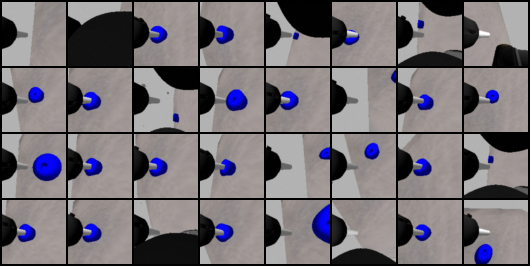}} \hspace{.01\columnwidth}
	\subfloat[Reconstructed successor samples at $n+1$]{\includegraphics[width=.45\columnwidth]{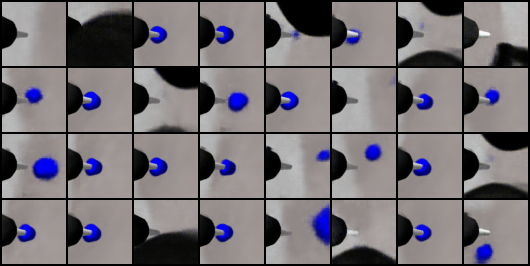}}
	\caption{Image samples/reconstructions (top) and successor samples/reconstructions (bottom) for latent dimension 16, KL=1e-4. At a sample rate of 20 Hz, the motion between $n$ and $n+1$ is small. \label{reconstructions}}
\end{figure}

The convolutional stack for the images increases from 32 to 128 filters, filter size 4, stride 2/1, with ReLU activations.  The force and proprioception encoders have one hidden layer of size 16, with a fusion encoder has a hidden layer of 64 units, fully-connected to mean and log standard deviation for $z$.  The decoder inverts this stack. Sample reconstructions can be seen in Figure \ref{reconstructions}.

The success classifier contains 2 FC layers followed by softmax (success labels are identified by hand). The next image predictor has an additional FC layer before the deconvolution stack. The next latent predictor uses a hidden layer of size 64, ReLU activations for mean and log standard deviation of $z_{n+1}$. 

The loss weights used for training are reconstruction $1e-4$, KL weight is $10$ normally, and $0.1$ with low KL encoders. The success weight is $0.1$, smoothing weight $500$, prediction of next $z$ weight $1$ and the prediction of next image weight $1e-4$. On the low KL smoothing, the smoothing weight was reduced to $0.1$.

\subsection{RL Agent}
A standard implementation of SAC is used \cite{yarats2020} and ablated to implement the DDPG agent. Actor and critic architecture have two hidden layers of size 96 or 128 (varied to average performance). The agents were trained with learning rates $0.003$ or $0.001$, replay buffer of 3e6, batch size of 256, and soft-update $\tau$ of $0.005$. Noise tuning was used on SAC, and DDPG had constant exploration noise of magnitude $0.25$ the maximum input. The agents were trained from 10 different random seeds each.

The simulation is done with the ODE dynamics engine \cite{smith2005}, under contact constraint settings of $CFM = 0.001$, $ERP = 0.2$ and $SOR = 1$, and maximum step size of $0.005$. The reward function largely follows that of \cite{lee2018a}.

\subsection{Derivation of bound on $\ln|J_g|$ \label{bnd_deriv}}
Recall that $\ln\left|J_{g}\right|=\ln\prod\left|\lambda_{i}\right|=\sum\ln\left|\lambda_{i}\right|$ where $\lambda_i,\,\,i\in[1,\dots,k]$ are the eigenvalues of $J_g\in\mathbb{R}^{k\times k}$. If $\delta_{z}\sim\mathcal{N}\left(0,\alpha J_{g}J_{g}^{T}\right)$, 

\begin{align*}
\ln E\left\Vert \delta_{z}\right\Vert _{2}^{2} & =\ln\mathrm{Tr}\left(\alpha J_{g}J_{g}^{T}\right)\\
& =\ln\sum\lambda_{i}^{2}+\ln\alpha.
\end{align*}

Applying the log sum inequality, where for positive constants $a_{i}$, $b_{i}$, $a=\sum a_{i}$, $b=\sum b_{i}$: 

\[
\sum a_{i}\ln\frac{a_{i}}{b_{i}}\geq a\ln\frac{a}{b}
\]
letting $a_{i}=1$, $b_{i}=\lambda_{i}^{2}$,
\begin{align*}
\sum\ln\frac{1}{\lambda_{i}^{2}} & \geq n\ln\frac{n}{\sum\lambda_{i}^{2}}\\
\sum\ln\lambda_{i}^{2} & \leq n\ln\sum\lambda_{i}^{2}-n\ln n\\
n^{-1}\sum\ln\lambda_{i}^{2}+\ln n & \leq\ln\sum\lambda_{i}^{2}
\end{align*}
By substitution,
\begin{align*}
\ln E\left\Vert \delta_{z}\right\Vert _{2}^{2} & \geq n^{-1}\sum\ln\lambda_{i}^{2}+\ln n+\ln\alpha\\
\Rightarrow & \frac{n}{2}\left(\ln E\left\Vert \delta_{z}\right\Vert _{2}^{2}-\ln\alpha n\right)\geq\ln\left|J_{g}\right|
\end{align*}

\end{document}